# One-class Collective Anomaly Detection based on LSTM-RNNs

Nga Nguyen Thi and Van Loi Cao
and Nhien-An Le-Khac

[1] Institute of Electronic, Institute of Military Science and Technology, Vietnam
[2] University College Dublin, Dublin, Ireland
*ngadtvt@gmail.com*, *loi.cao@ucdconnect.ie*
*an.lekhac@ucd.ie*

**Abstract.** Intrusion detection for computer network systems has been becoming one of the most critical tasks for network administrators today. It has an important role for organizations, governments and our society due to the valuable resources hosted on computer networks. Traditional misuse detection strategies are unable to detect new and unknown intrusion types. In contrast, anomaly detection in network security aims to distinguish between illegal or malicious events and normal behavior of network systems. Anomaly detection can be considered as a classification problem where it builds models of normal network behavior, of which it uses to detect new patterns that significantly deviate from the model. Most of the current approaches on anomaly detection is based on the learning of normal behavior and anomalous actions. They do not include memory that is they do not take into account previous events classify new ones. In this paper, we propose a one-class collective anomaly detection model based on neural network learning. Normally a Long Short-Term Memory Recurrent Neural Network (LSTM RNN) is trained only on normal data, and it is capable of predicting several time-steps ahead of an input. In our approach, a LSTM RNN is trained on normal time series data before performing a prediction for each time-step. Instead of considering each time-step separately, the observation of prediction errors from a certain number of time-steps is now proposed as a new idea for detecting collective anomalies. The prediction errors of a certain number of the latest time-steps above a threshold will indicate a collective anomaly. The model is evaluated on a time series version of the KDD 1999 dataset. The experiments demonstrate that the proposed model is capable to detect collective anomaly efficiently.

## 1  Introduction

Network anomaly detection refers to the problem of detecting illegal or malicious activities or events from normal connections or expected behavior of network systems [3, 4]. It has become one of the most popular subjects in the network security domain due to the fact that many organizations and governments are now seeking good solutions to protect valuable resources on computer networks from unauthorized and illegal accesses, network attacks or malware. Over the last three decades, machine learning techniques are known as a common approach for developing network anomaly detection models [2, 3]. Network anomaly detection is usually posed as a type of classification problem: given a dataset representing normal and anomalous examples, the goal is to build a learning classifier which is capable of signaling when a new anomalous data sample is encountered [4].

Most of the existing approaches consider an anomaly as a discrete single data point: cases when they occur "individually" and "separately" [5, 6, 15]. In such approaches, anomaly detection models do not have the ability to represent the information from previous data points or events for evaluating a current point. In network security domain however, some kinds of attacks (e.g. Denial of Service - DoS) usually occur for a long period of time (several minutes) [9], and are often represented by a sequence of single data points. Thus these attacks will be indicated only if a sequence of single data points are considered as attacks. In this context, network data can be considered as a time series data that are sequences of events obtained over repeated measurements of time. Many approaches ranging from statistical techniques to machine learning techniques are employed for analyzing time series data, and their efficiency has been proven in time series forecasting problems. These approaches are based on the information of previous events to forecast the incoming step. In order to detect these kinds of attack mentioned above, anomaly detection models should be capable of memorizing the information from a number of previous events, and representing the relationship between them and with the current event. The model should not only have the ability to estimate prediction errors anomalous scores for each individual time-step but also be able to observe sequences of time-steps that are potential to be collective anomaly. To avoid important mistakes, one must always consider every outcome: in this sense a highly anomalous value may still be linked to a perfectly normal condition, and conversely. In this work, we aim to build an anomaly detection model for this kind of attacks (known as collective anomaly detection in [4]).

Collective anomaly is the term to refer to a collection of related anomalous data instances with respect to the whole dataset [4]. A single data point in a collective anomaly may not be considered as anomalies by itself, but the occurrence of a sequence of single points together may indicate a collective anomaly. Hidden Markov model, Probabilistic Suffix Trees, etc. are popular techniques for collective anomaly detection [4]. Recently, Long Short-Term Memory Recurrent Neural Network [7] has been recognized as a powerful technique to represent the relationship between a current event and previous events, and handles time series problems [11, 13]. However, these approaches are proposed only for predicting time series anomalies at individual level (predicting prediction error anomalous score for each time-step), not at the collective level (observing prediction errors for a sequence of time-steps). Moreover, both normal and anomalous data are employed for training stage: either for training process (constructing classified models) or validation process (estimating model parameters). Thus, such the models are limited to detect new kinds of network attack. Collecting and labeling anomalous data are also expensive and time-consuming tasks. Therefore, we will propose a collective anomaly detection model by using the predictive power of LSTM RNN. The ability to detect collective anomaly of the proposed model will be demonstrated on DoS attack group in the KDD Cup 1999 dataset.

The rest of the paper is organized as follows. We briefly review some work related to anomaly detection and LSTM RNN in Section 2. In Section 3, we give a short introduction to LSTM RNN. This is followed by a section proposing the collective anomaly detection model using LSTM RNN. Experiments, Results and Discussion are presented in Section 5 and Section 6 respectively. The paper concludes with highlights and future directions in Section 7.

## 2   Related Work

When considering a time series dataset, point anomalies are often directly linked to the value of the considered sample. However, attempting real time collective anomaly detection implies always being aware of previous samples, and more precisely their behavior. This means that every time-step should include an evaluation of the current value combined with the evaluation of preceding information. In this section, we briefly describe previous work applying LSTM RNN for time series and collective anomaly detection problems [11, 13, 14, 16].

Olsson et al. [14] proposed an unsupervised approach for detecting collective anomalies. In order to detect a group of the anomalous examples, the anomalous score of the group of data points was probabilistically aggregated from the contribution of each individual example. Obtaining the collective anomalous score was carried out in an unsupervised manner, thus it is suitable for both unsupervised and supervised approaches to scoring individual anomalies. The model was evaluated on an artificial dataset and two industrial datasets, detecting anomalies in moving cranes and anomalies in fuel consumption.

In [11], Malhotra et al. applied a LSTM network for addressing anomaly detection problem in time series fashion. A stacked LSTM network trained on only normal data was used to predict values of a number of time-steps (L steps) ahead. Thus, the prediction model produced L prediction values in the period of L time-steps for a single data point. It then resulted in a prediction error vector with L elements for each data point. The prediction error of a single point was then computed by modeling its prediction error vector to fit a multivariate Gaussian distribution, which was used to assess the likelihood of anomaly behavior. Their model was demonstrated to perform well on four datasets.

Marchi et al. [13, 12] presented a novel approach by combining non-linear predictive denoising autoencoders (DA) with LSTM for identifying abnormal acoustic signals. Firstly, LSTM Recurrent DA was employed to predict auditory spectral features of the next short-term frame from its previous frames. The network trained on normal acoustic recorders tends to behave well on normal data, and yields small reconstruction errors whereas the reconstruction errors from abnormal acoustic signals are high. The reconstruction errors of the autoencoder was used as an "anomaly score", and a reconstruction error above a predetermined threshold indicates a novel acoustic event. The model was trained on a public dataset containing in-home sound events, and evaluated on a dataset including new anomaly events. The results demonstrated that their model performed significantly better than existing methods. The idea is also used in a practical acoustic example [13, 12], where LSTM RNNs are used to predict short-term frames.

In [16] Ralf C et at. employed LSTM-RNN for intrusion detection problem in supervised manner. The processed version of the KDD Cup 1999 dataset, which is represented in time-series, were fed to the LSTM-RNN. The network has five outputs representing the four groups of attacks and normal connections in the data. Both labeled normal connections and labeled attacks were used for training the model and estimating the best LSTM-RNN architecture and its parameters. The selected model was then evaluated on 10% of the corrected dataset under measurements of confusion matrix and accuracy. The results shown that their model performed well in terms of accuracy, especially it achieved very high performance on two groups of attacks, Probe and DoS.

To the best of our knowledge, non of previous work using LSTM RNN addresses the problem of collective anomaly detection. Thus, we aim to develop collective anomaly detection using LSTM RNN. Our solution consists of two stages: (1) LSTM RNN will be employed to represent the relationship between previous time-steps and current one in order to estimate anomalous score (known as prediction error) for each time-step. This stage is considered as developing a time series anomaly detection, and similar to the previous work [11]; (2) A method will be proposed for observing sequences of single data points based on their anomalous scores to detect collective anomaly. The second stage makes our work original and different from previous work that applied LSRM RNN for time series anomaly detection. This will prove very efficient in our example: First, we will train an LSTM RNN on only normal data in order to learn the normal behavior. The trained model will be validated on normal validation set in order to estimate the model parameters. The result classifier will then employ to rate the anomalous score for data at each time-step. The anomalous score of a sequence of time steps will be aggregated from the contribution of each individual one. By imposing a predetermine threshold a sequence of single time-steps will indicate as a collective anomaly if its anomalous score is higher than the threshold. More details on our approach can be found in Section 4.

## 3    Preliminaries

In this section we briefly describe the structure of Long Short Term Memory nodes, and the architecture of a LSTM RNN using LSTM hidden layer. The LSTM was proposed by Hochreiter et al. [7] in 1997, and has already proven to be a powerful technique for addressing the problem of time series prediction.

The difference initiated by LSTM regarding other types of RNN resides in its "smart" nodes presented in Hidden layer block in Fig. 1. Each of these cells contains three gates, input gate, forget gate and output gate, which decide how to react to an input. Depending on the strength of the information each node receives, it will decide to block it or pass it on. The information is also filtered with the set of weights associated with the cells when it is transferred through these cells.

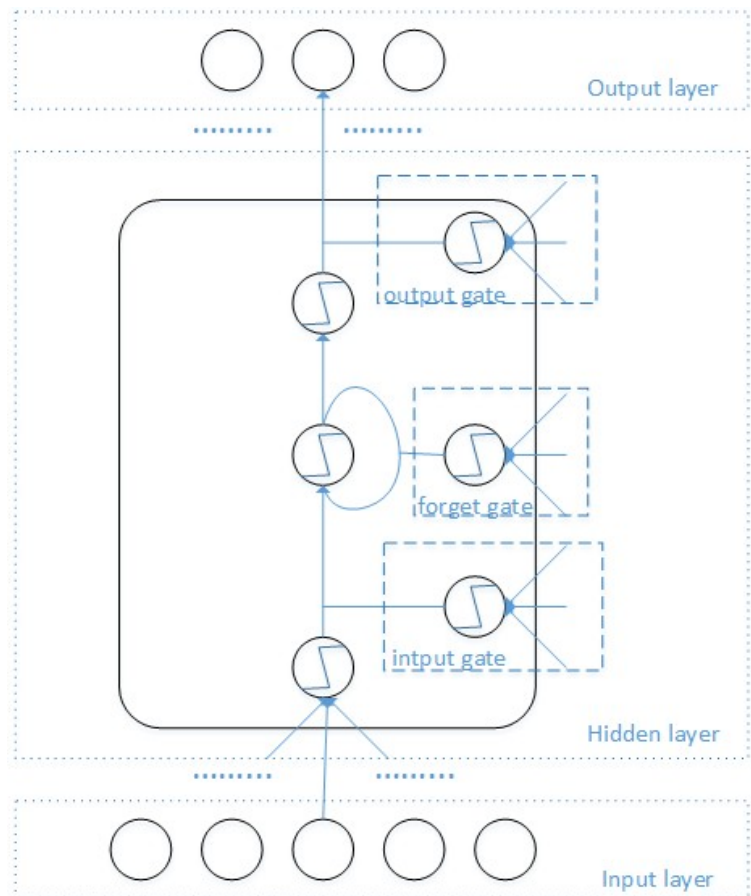

**Fig. 1.** LSTM RNN Architecture

The LSTM node structure enables a phenomenon called backpropagation through time. By calculating for each hidden layer the partial derivatives of the output, weight and input values, the system can move backwards to trace the evolving error between real output and predicted output. Afterwards, the network uses the derivative of this evolution to adapt its weights and decrease prediction error. This learning method is named Gradient descent.

A simple LSTM RNN as in Fig 1 consists three layers: Input layer, LSTM hidden layer and output layer. The input and output layers are the same as those in multi-layered perceptrons (MLPs). The input nodes are the input data, and the output nodes can be sigmoid, tanh or other transform functions. The LSTM hidden layer is formed from a number of the "smart" nodes that are fully connected to the input and output nodes. Two common techniques, Gradient descent and Back-propagation can be used for optimizing its loss function and updating its parameters.

As mentioned before, Long Short-Term Memory has the power to incorporate a behaviour into a network by training it with normal data. The system becomes representative of the variations of the data. In other words, a prediction is made focusing on two features: the value of a sample and its position at a specific time. This means that two same input value at different times probably results in two different outputs. It is because a LSTM RNN is stateful, i.e. has a "memory", which changes in response to inputs.

## 4  Proposed Approach

As mentioned in the Related work section, one of the most recent research using LSTM RNNs for building anomaly detection model in time series data is from [11]. The model was demonstrated to be efficient for detect anomalies of time series data. In their model, the prediction errors of a data point is computed by fitting its prediction errors vector to a multivariate Gaussian distribution which is then used to assess the likelihood of anomalous behavior. Hence, it is only suitable for detecting abnormal events that happen instantly i.e. in a very short time such as in the electrocardiogram or power demand applications since the model do not have the ability of estimating likelihood anomaly for a long period of time (a sequence of data points). Consequently, the model is not suitable for the collective anomaly detection in the context of network security where some kinds of network attack last for a significant period time.

Therefore, in this paper we propose a new approach of using LSTM RNNs for cyber-security attacks at collective level. We will use a simple LSTM RNN architecture, in contrast to a stacked LSTM in [11]. This does not change the core principle of the method: when given sufficient training, a LSTM RNN adapts its weights, which become characteristic of the training data. In the network, each output nodes represents each time-step that the model predicts. For example, if the model is fed a data point at time-step t-th for predicting values at next three steps ahead, the first, second and third output nodes will represent the values at time-steps $(t + 1)$-th, $(t + 2)$-th and $(t + 3)$-th respectively. Instead of fitting a mixture Gaussian model, we simple compute the prediction error of a individual data point by mean over the prediction errors at three time-steps. We show the LSTM RNNs ability to learn the behavior of a training set, and in this stage it acts like a time series anomaly detection model. Following this stage, we proposed terms (defined below) to monitor the prediction errors of a certain number of successive data points and locate collective anomaly in data. The second stage allows our model to detect collective anomaly, and make model original and different from previous ones. Thus, the performance of our model is not compared to that of previous models using LSTM RNN on time series anomaly detection, e.g. [11].

Terms for measuring prediction errors of data point, and monitoring anomalous behavior in a period of time-steps are defined as below:

- **Relative Error (RE):** the relative prediction error between a real value $x$ and its prediction value $\hat{x}$ from LSTM RNN at every time-step is defined as in eq.1. Note that a single data can not be considered as a collective anomaly. However, the larger value of RE a data point has, the higher probability the single data point belongs to a collective anomaly.

$$RE(x, \hat{x}) = |x - \hat{x}| \qquad (1)$$

- **Prediction Error Threshold (PET):** It is employed to determiner an individual query point (a time-step) can be classified as a normal time-step or considered as an element in a potential collective anomaly. Its prediction error RE above threshold PET may indicate a element in a collective anomaly.
- **Collective Range (CR):** a minimum number of anomalies appearing successively in a network flow are considered as a collective anomaly. Both PET and CR are estimated based on the best classification performance of the model on normal validation set.

## 5 Experiments

### 5.1 Datasets

In order to demonstrate the efficient performance of the proposed model, we choose a dataset related to the network security domain, the KDD 1999 dataset [1, 8], for our experiments. The dataset in tcpdump format was collected from a simulated military-like environment over a period of 5 weeks (from $1^{st}$ March 1999 to $9^{th}$ April 1999). The dataset is composed of a two-weeks for training, weeks 1 and week 3 (free attack), one week for validation, week 2 (labeled attacks), and other two weeks for testing, weeks 4 and 5 (both normal and anomalous data).

There are four main groups of attacks in the dataset, but we restrict our experiments on a specific attack, Neptune, in the Denial-of-Service group. The dataset is also converted into a time series version before feeding into these models. More details about how to obtain a time series version from the original tcpdump dataset, and how to choose training, validation and testing sets are presented in the following paragraphs.

The first crucial step is to build a conveniently usable time series dataset out of the tcpdump data, and to select the interested features. We use terminal commands and a python program to convert the original tcpdump data in the KDD 1999 dataset into a time dependent function. This method is a development of the proposed transformation in [10] that acts directly on the tcpdump to obtain real time statistics of the data. Our scheme follows this step by step transition as described below:

$$\text{tcpdump} \Rightarrow \text{pcap} \Rightarrow \text{csv}$$

Each day of records can be time-filtered and input into a new .pcap file. This also has the advantage of giving a first approach on visualizing the data by using Wireshark functionalities. Once this is done, the tshark command is adapted to select and transfer the relevant information from the records into a csv format. We may note that doing this is a first step towards faster computation and better system efficiency, since all irrelevant pcap columns can be ignored. Although the method for converting data do not suit for the model performing real-time, it is sufficient enough for evaluating the proposed model in detecting collective anomaly as the main objective of this work. However, if the attack data is recorded in real-time under time series format, then our method can be applied in real-time detection.

In our experiments, we use 6-days normal traffic in the first and third weeks for training, $n_{train}$, and one-day normal traffic (Thursday) in week 3 for validation, $n_{valid}$. Testing sets include 1-day normal traffic from week 3 (Friday), $n_{test}$, and 1-day data containing attacks in week 2 (Wednesday), $a_{test}$. The protocol will be the following: training the network with $n_{train}$, using $n_{valid}$ for choosing Prediction Error Threshold (PET) and Collective Range (CR), and evaluating the proposed models on $n_{test}$ and $a_{test}$.

**Table 1.** Parameter Settings

| LSTM RNN Parameters | |
|---|---|
| Input Size | 1 |
| Hidden LSTM Layer | 10 |
| Output Sigmoid Layer | 1, 2 or 3 |
| Learning Rate | $10^{-4}$ |
| Number of Epochs | 100 |
| Momentum | 0.5 |
| Batch Size | 1 |
| **Collective Thresholds** | |
| Prediction Error Threshold (PET) | 0.3 |
| Collective Range (CR) | 4 |

### 5.2 Experimental Settings

Our experiments are aim to demonstrate the ability of detecting collective anomalies of the proposed model. However, there is no anomalous instances available for training and validation stages, it is much harder than binary classification problem in estimating hyper-parameters, optimizing network architectures and setting thresholds. Thus, we will briefly discus this issues in next paragraph, and design two experiments, a preliminary experiment for choosing these thresholds and a main experiment for evaluating the proposed model.

We investigate three network architectures. The difference amongst these architectures is only the size of output layer, with one, two and three outputs for predicting 1-step, 2-step and 3-step ahead respectively. This means that the three-output network can predict three values for three steps ahead with a current input. The number of hidden nodes and the learning rate can strongly influence the performance of a LSTM RNN. Each synapse of a network is weighted differently, and can be considered as a unique interpretation of the input data. Each node of the hidden layer is storage space for these interpretations. Theoretically, the higher number of hidden nodes, the more information the network can contain. This also means more computation, and may lead to over-fitting. Therefore, it should be trade off between the detection rate and the computational expenses on constructing models and querying new data. In this paper, we choose the size of LSTM hidden layer equal to 10. The learning rate is another factor directly linked to the speed at which a LSTM RNN can improve its predictions. For a time step t during training process, the synapse weights of the neural network are updated. The learning rate defines how much we wish a weight to be modified at each instant. We choose a common used value for learning rate, $10^{-4}$.

The preliminary experiment aim to tune Prediction Error Threshold (PET) and Collective Range (CR) by using only normal validation set, $n_{valid}$. This means that these parameters are choose so as to the model can correctly classify most of instances in $n_{valid}$ (say 95%, 97% or 100%) belonging to normal class. In this paper, we set these threshold so as to keep 100% of examples in $n_{valid}$ as normal data. The choice of CR depends on how long a single data point represents for. In our data, each single point represents a period of 10 minutes, so CR. Thus we choose CR equal to 4 which is equivalent to a period of 40 minutes. One CR chosen, we will compute detection rate of the model with 20 different values of PET ranging from 0.05 to 1.0. on $n_{valid}$. The smallest value of PET that enables the model to correctly classify 100% examples in $n_{valid}$ is chosen. More details about network architectures, parameters settings are presented in Table 1.

The main experiment is to shown the ability of LSTM-RNNs in detecting a disproportionate durable change in a time series anomaly. Once the preliminary experiment is complete, these trained models with collective thresholds, PET and CR is employed to detect anomalous region in data, $n_{valid}$, $n_{test}$ and $a_{test}$. This experimental results include the prediction error of each single data point, specific anomalous regions on normal validation set and test sets from the three models. The training error is plotted in Fig 2. Fig 3, 4 and 5 illustrate the prediction errors on validation set ($v_n$), testing sets ($n_{test}$ and $a_{test}$). Specific regions predicted as collective anomaly and the proportion of these anomalous regions are presented in Table 2.

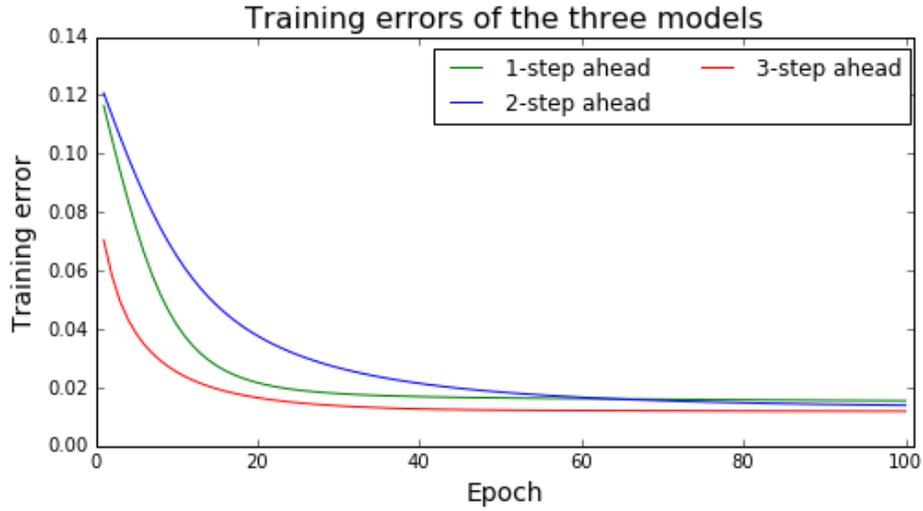

**Fig. 2.** The training errors of the proposed model

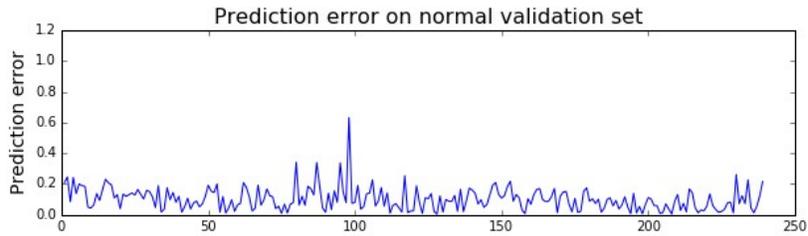

**Fig. 3.** The prediction error from 3-step ahead model on validation set.

## 6    Results and Discussion

The Table 2 shows the collective anomaly prediction of the proposed model on three datasets, $n_{valid}$, $n_{test}$ and $a_{test}$. The collective anomaly prediction includes specific regions in data and the percentage of data instances (time-steps) within these regions. There is no anomalous region found in normal validation set, $n_{valid}$ because the thresholds, $PET$ and $CR$ have been tuned to classify 100% of $n_{valid}$ belonging to normal class. Only one regions in normal test, $n_{test}$ is found by the third classifier (3-step ahead), which is False Negative. In anomaly test, many regions are detected as collective anomalies by both of three classifiers. It seems to be that the more steps ahead the model can predict, the more regions can be found. They are three regions (8.79%), five regions (12.13%) and six regions (14.23%) found by the first, second and third classifiers respectively.

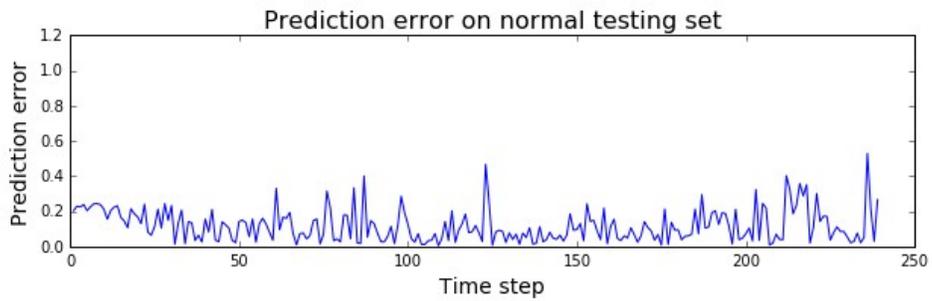

**Fig. 4.** The prediction error from 3-step ahead model on normal test.

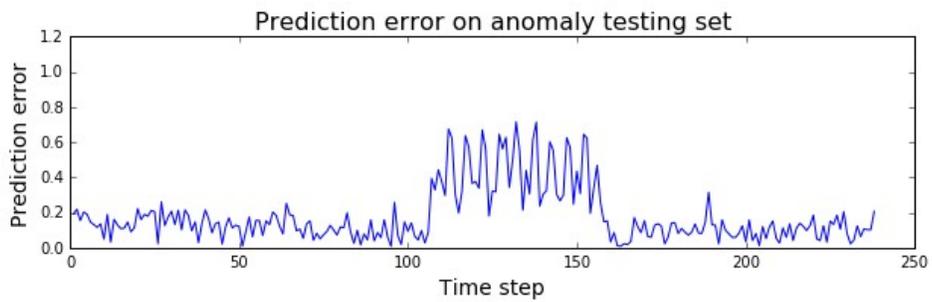

**Fig. 5.** The prediction error from 3-step ahead model on anomaly test.

**Table 2.** The prediction collective anomalies in validation and test sets

| Dataset | 1-Step Ahead | | 2-Step Ahead | | 3-Step Ahead | |
|---|---|---|---|---|---|---|
| | Anomaly region | Anomaly ratio | Anomaly region | Anomaly ratio | Anomaly region | Anomaly ratio |
| $n_{valid}$ | - | 0.0% | - | 0.0% | - | 0.0% |
| $n_{test}$ | - | 0.0% | - | 0.0% | 215 - 219 | 1.67% |
| $a_{test}$ | 107 - 111<br>116 - 124<br>125 - 134<br>-<br>-<br>- | 8.79% | 107 - 111<br>116 - 124<br>125 - 134<br>135 - 139<br>140 - 144<br>- | 12.13% | 107 - 111<br>116 - 124<br>125 - 134<br>135 - 139<br>140 - 145<br>150 - 154 | 14.23% |

The fig 2 illustrates the training errors from three classifiers with 1-step, 2-step and 3-step prediction ahead. These errors tend to converge after 100 epochs, and the error curve of the 3-step classifier levels off earlier than two others. The fig 3, 4 and 5 also present the prediction errors of the third classifiers on $n_{valid}$, $n_{test}$ and $a_{test}$. In fig 3 and 4, the prediction errors are just fluctuated around 0.2, and few individual errors have large values. Thus, these datasets are not considered as collective anomaly. However, the error patterns in fig 5 are quite different. The errors within time steps 110 to 155 is quite high, much higher than the rest of error regions. This regions are detected as collective anomalies as presented in Table 2.

In training stage, the more prediction time-steps a model has, the less training error the model produces (see fig. 2). This suggests that the models with more prediction time-steps tend to learn the normal behaviors of network traffic more efficiently than ones with less prediction time-steps. However, the number of prediction time-steps can enrich the model's ability to detect anomaly regions in the anomaly testing set $a_{test}$. This may imply that the three-steps model is more robust to learn the normal behaviors and identify collective anomalies than the two others.

## 7 Conclusion and Further work

In this paper, we have proposed a model for collective anomaly detection based on Long Short-Term Memory Recurrent Neural Network. We have motivated this method through investigating LSTM RNN in the problem of time series, and adapted it to detect collective anomalies by proposing the measurements in Section 4. We investigated the hyper-parameters, the suitable number of inputs and some thresholds by using the validation set.

The proposed model is evaluated by using the time series version of the KDD 1999 dataset. The results suggest that proposed model is efficiently capable of detecting collective anomalies in the dataset. However, they must be used with caution. The training data fed into a network must be organized in a coherent manner to guarantee the stability of the system. In future work, we will focus on how to improve the classification accuracy of the model. We also observed that implementing variations in a LSTM RNNs number of inputs might trigger different output reactions.

## References

1. DARPA intrusion detection evaluation. (n.d.). (Retrieved June 30, 2016), *http://www.ll.mit.edu/ideval/data/1999data.html*

2. Ahmed, M., Mahmood, A.N., Hu, J.: A survey of network anomaly detection techniques. Journal of Network and Computer Applications 60, 19–31 (2016)
3. Bhattacharyya, D.K., Kalita, J.K.: Network anomaly detection: A machine learning perspective. CRC Press (2013)
4. Chandola, V., Banerjee, A., Kumar, V.: Anomaly detection: A survey. ACM computing surveys (CSUR) 41(3), 15 (2009)
5. Chmielewski, A., Wierzchon, S.T.: V-detector algorithm with tree-based structures. In: Proc. of the International Multiconference on Computer Science and Information Technology, Wisła (Poland). pp. 9–14. Citeseer (2006)
6. Hawkins, S., He, H., Williams, G., Baxter, R.: Outlier detection using replicator neural networks. In: International Conference on Data Warehousing and Knowledge Discovery. pp. 170–180. Springer (2002)
7. Hochreiter, S., Schmidhuber, J.: Long short-term memory. Neural computation 9(8), 1735–1780 (1997)
8. KDD Cup Dataset (1999), available at the following website *http://kdd.ics.uci.edu/databases/kddcup99/kddcup99.html*
9. Lee, W., Stolfo, S.J.: A framework for constructing features and models for intrusion detection systems. ACM transactions on Information and system security (TiSSEC) 3(4), 227–261 (2000)
10. Lu, W., Ghorbani, A.A.: Network anomaly detection based on wavelet analysis. EURASIP Journal on Advances in Signal Processing 2009, 4 (2009)
11. Malhotra, P., Vig, L., Shroff, G., Agarwal, P.: Long short term memory networks for anomaly detection in time series. In: Proceedings. p. 89. Presses universitaires de Louvain (2015)
12. Marchi, E., Vesperini, F., Eyben, F., Squartini, S., Schuller, B.: A novel approach for automatic acoustic novelty detection using a denoising autoencoder with bidirectional lstm neural networks. In: 2015 IEEE International Conference on Acoustics, Speech and Signal Processing (ICASSP). pp. 1996–2000. IEEE (2015)
13. Marchi, E., Vesperini, F., Weninger, F., Eyben, F., Squartini, S., Schuller, B.: Non-linear prediction with lstm recurrent neural networks for acoustic novelty detection. In: 2015 International Joint Conference on Neural Networks (IJCNN). pp. 1–7. IEEE (2015)
14. Olsson, T., Holst, A.: A probabilistic approach to aggregating anomalies for unsupervised anomaly detection with industrial applications. In: FLAIRS Conference. pp. 434–439 (2015)
15. Salama, M.A., Eid, H.F., Ramadan, R.A., Darwish, A., Hassanien, A.E.: Hybrid intelligent intrusion detection scheme. In: Soft computing in industrial applications, pp. 293–303. Springer (2011)
16. Staudemeyer, R.C., Omlin, C.W.: Evaluating performance of long short-term memory recurrent neural networks on intrusion detection data. In: Proceedings of the South African Institute for Computer Scientists and Information Technologists Conference. pp. 218–224. ACM (2013)